\documentclass[10pt,twocolumn,letterpaper]{article}

\usepackage[pagenumbers]{cvpr} %

\newcommand{\figref}[1]{Fig.~\ref{#1}}
\newcommand{\secref}[1]{Section~\ref{#1}}

\newcommand{\tabref}[1]{Table~\ref{#1}}

\makeatletter
\DeclareRobustCommand\onedot{\futurelet\@let@token\@onedot}
\def\@onedot{\ifx\@let@token.\else.\fi}
\def\eg{e.g\onedot} 
\def\ie{i.e\onedot}

\makeatother

\definecolor{darkgreen}{rgb}{0,0.7,0}
\definecolor{darkblue}{RGB}{31,119,180}
\definecolor{darkred}{RGB}{214,39,40}
\definecolor{mediumgray}{rgb}{0.5,0.5,0.5}
\definecolor{mediumteal}{rgb}{0,0.5,0.5}
\definecolor{naviblue}{RGB}{0,0,128}

\definecolor{amber}{HTML}{FFBE28}
\definecolor{flame}{HTML}{E65100}
\definecolor{indigo}{HTML}{535ABF}
\definecolor{cyan}{HTML}{40C1D1}

\definecolor{teal}{HTML}{40888B}
\definecolor{fern}{HTML}{81C784}
\definecolor{mint}{HTML}{E8F5E9}
\definecolor{slate}{HTML}{263238}

\definecolor{ellisred}{rgb}{0.87,0.44,0.38} %
\definecolor{ellisgreen}{rgb}{0.69,0.90,0.52} %
\definecolor{elliscyan}{rgb}{0.29,0.77,0.74} %
\definecolor{ellisorange}{rgb}{0.89,0.55,0.28} %
\definecolor{ellisblue}{rgb}{0.41,0.61,0.86} %

\definecolor{customgray}{RGB}{136, 138, 133}

\newcommand{\boldparagraph}[1]{\vspace{0.1cm}\noindent{\bf #1:} }

\definecolor{darkgreen}{rgb}{0,0.7,0}
\definecolor{lgray}{rgb}{0.6,0.6,0.6}

\definecolor{mediumgray}{rgb}{0.5,0.5,0.5}

\definecolor{cvprblue}{rgb}{0.21,0.49,0.74}
\usepackage[pagebackref,breaklinks,colorlinks,allcolors=cvprblue]{hyperref}

\usepackage{xcolor}
\usepackage[normalem]{ulem}  %

\usepackage{algorithm}
\usepackage{algpseudocode}
\definecolor{sourcolor}{HTML}{76B900}
\newcommand{\sour}[1]{\textcolor{sourcolor}{#1}}
\usepackage{colortbl}

\def\confName{CVPR}
\def\confYear{2027}

\title{Pisco Sour: Blending Distillation and Demonstrations for End-to-End Driving}

\newcommand{\authorrow}[1]{{\fontsize{11}{13.6}\selectfont #1}}
\author{
\authorrow{Daniel Dauner$^{1,2}$\thanks{Work performed in part during an internship at NVIDIA.}\hspace{0.4em}\quad Maximilian Igl$^{1}$\quad Peter Karkus$^{1}$\quad Kashyap Chitta$^{2,3}$\quad Lan Feng$^{1,4}$\quad Yuxiao Chen$^{1}$}\\
\authorrow{Alexander Naumann$^{1}$\quad Jef Packer$^{1}$\quad Andreas Geiger$^{2,5}$\quad Boris Ivanovic$^{1}$\quad Marco Pavone$^{1,6}$}\\[6pt]
{\normalsize $^1$NVIDIA \quad $^2$KE:SAI \quad $^3$ELLIS Institute T\"ubingen \quad $^4$EPFL}\\[-2pt]
{\normalsize $^5$University of T\"ubingen, T\"ubingen AI Center \quad $^6$Stanford University}\\[5pt]
{\normalsize \href{https://danieldauner.github.io/pisco-sour}{\texttt{danieldauner.github.io/pisco-sour}}}
}
\begin{document}
\maketitle
\begin{abstract}

End-to-end driving policies that imitate human driving logs are limited by the state coverage of the recordings, which makes these models prone to compounding errors in closed-loop deployment.
Addressing this via closed-loop training, such as reinforcement learning (RL), remains challenging.
RL for end-to-end driving requires exploration in costly sensor simulations to optimize a reward function that specifies only a subset of the desired qualities of driving behavior.
In this work, we propose {Pisco Sour}, a post-training recipe for end-to-end driving models that mixes on-policy distillation of a student policy from a privileged RL teacher (Pisco) with human demonstrations (Sour).
The teacher is trained with RL in lightweight simulations of bounding boxes and maps, while our distillation in a sensor simulator uses about $700\times$ fewer environment interactions than training the teacher.
Regularizing the distillation with human demonstrations reduces undesired behaviors that the student would otherwise inherit from the teacher's limited observation space and reward function.
In AlpaSim, Pisco Sour raises the distance driven per infraction of a 68M-parameter TransFuser from 1.7 to 3.1 km and of the 10B-parameter vision-language-action model FlashDrive from 0.9 to 5.6 km. Pisco gives most of the score gain, and Sour restores human driving norms at equal or higher score.
\end{abstract}

\section{Introduction}
\label{sec:intro}

End-to-end (E2E) learning has become the dominant paradigm in autonomous driving~\cite{Chitta2023PAMI, Hu2023CVPR, Wang2025ARXIV,Chen2023PAMI}, where a single network maps raw sensor data directly to driving outputs.
Human demonstrations provide easy-to-collect labels of safe and natural driving behavior for behavior cloning (BC)~\cite{Pomerleau1988NIPS,Codevilla2018ICRA}.
However, this supervision is limited to the demonstrator's own state distribution, and the resulting distribution shift leads to compounding errors and poor robustness once the policy drives in closed loop~\cite{Ross2010AISTATS, Codevilla2019ICCV, Dauner2023CORL}.

\begin{figure}
    \centering
    \includegraphics[width=1.0\linewidth]{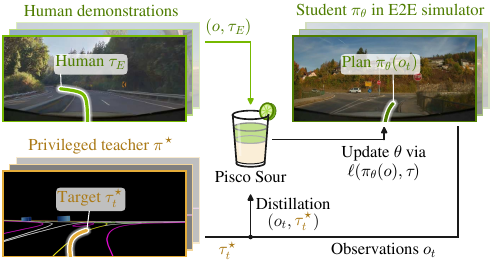}
    \vspace{-0.6cm}
    \caption{\textbf{Pisco Sour.}
    An end-to-end student policy~$\pi_\theta$ is rolled out in simulation to generate planned trajectories from visual observations~$o_t$ (top right).
    At every step, a privileged teacher $\pi^\star$ provides a target trajectory $\tau^\star_t$ for the student's current state, based on privileged map and bounding box annotations.
    \textbf{Pisco:} The teacher's targets are distilled into the student using the on-policy pairs $(o_t, \tau^\star_t)$ (bottom left). \textbf{Sour:} Each update also mixes in human demonstrations $(o, \tau_E)$ with the same loss $\ell$ (top left).}
    \label{fig:teaser}
\end{figure}

Training on closed-loop rollouts of the policy itself can mitigate this distribution shift, but has two requirements. First, such training requires a simulator, since training on public roads is dangerous. Advances in neural reconstruction~\cite{Mildenhall2020ECCV,Kerbl2023ACM,Tonderski2024CVPR,Chen2025ICLR} have enabled E2E simulators, such as AlpaSim~\cite{Nvidia2025ALPASIM} and HUGSIM~\cite{Zhou2025PAMI}, that provide photorealistic observations of states the recording vehicle has not visited. Second, closed-loop training requires a feedback signal at every step, which existing methods formalize differently.

In reinforcement learning (RL)~\cite{Sutton1998MIT,Jaeger2024FTO}, the policy is optimized through large-scale exploration to maximize a reward function.
On high-definition (HD) map and bounding box observations, RL has yielded robust privileged planners that outperform rule-based planners in closed-loop driving~\cite{Cusumano2025ICML, Jaeger2025CORL}.
For sensor-based policies, such as vision-language-action (VLA) models~\cite{Wang2025ARXIV, Nvidia2026ALPAMAYO2SUPER, Zhou2026ARXIV}, every step requires running a large network and rendering camera images in an E2E simulator, which makes exploration at scale prohibitively expensive~\cite{Cusumano2025ICML,Da2026ARXIV}.
In addition, a reward function for real-world driving is hard to design~\cite{Knox2023Elsevier} and specifies only a subset of the desired driving behavior, e.g. progress and safety based on scene annotations~\cite{Gulino2023NEURIPS, Jaeger2025CORL, Kazemkhani2025ICLR}. 

In contrast, methods like DAgger~\cite{Ross2011AISTATS} or on-policy distillation~\cite{Agarwal2024ICLR,Da2026ARXIV} roll out the policy and query an expert for a label at every visited state, thereby training with a supervised objective directly on the policy's own state distribution.
While less data intensive than RL, this requires a scalable source of labels for arbitrary states, which human annotation cannot provide.
RoaD~\cite{Garcia2026CVPR} avoids the expert by guiding the rollouts towards the human log, which restricts the supervision to states close to the recorded drive.
Privileged RL planners~\cite{Da2026ARXIV} can provide such expert labels at any state. The expensive exploration and training of RL can take place in lightweight simulators that only model boxes and maps. 
The trained teacher can then be queried on states a student visits to gather recovery trajectories, as shown in~\figref{fig:teaser}.

We argue that RL teachers alone cannot replace behavior cloning of human driving.
Human demonstrations reflect safe, comfortable driving based on social norms, though their supervision signal is fixed to states visited during data collection.
RL teachers can provide trajectory labels at arbitrary states, but perceive only an abstract representation of a scene, and act based on an engineered reward function used during training~\cite{Cusumano2025ICML, Jaeger2025CORL,Zhang2021ICCV}. Indeed, we find that students trained to imitate RL teachers inherit weaknesses of the teacher's inputs and reward design, such as running red lights or tailgating, and effectively forget behaviors learned during pre-training on human demonstrations. In prior work on RL teacher distillation, human demonstrations either serve only as a pre-training prior~\cite{Da2026ARXIV} or are omitted entirely for pure teacher-only supervision~\cite{Zhang2021ICCV,Rowe2026CORL,Xiong2026ARXIV,He2026ARXIV}.

In this paper, we propose a training recipe consisting of two parts (\figref{fig:teaser}).
Pisco (\underline{P}olicy \underline{i}mitation through \underline{s}tudent-\underline{co}ntrolled rollouts) distills a privileged RL teacher into a pre-trained end-to-end student on the student's own state distribution to learn recovery drives for closed-loop robustness. 
Sour (\underline{S}upervision from \underline{o}pen-loop h\underline{u}man \underline{r}ecordings) mixes human demonstrations during the distillation training, counteracting the limitations in the teacher's observation space and reward design. 

Our contributions are: 
\textbf{(1)} Pisco Sour, a post-training strategy for any end-to-end driving model with trajectory predictions in closed-loop simulation, where we mix on-policy distillation from RL teachers with human supervision. Our method applies to both a 68M-parameter TransFuser~\cite{Chitta2023PAMI, Nguyen2026CVPR} and the 10B-parameter VLA FlashDrive~\cite{Li2026CORL,Wang2025ARXIV}, and raises the distance driven per infraction from 1.7 to 3.1 km and from 0.9 to 5.6 km, respectively.
\textbf{(2)} A comparison of RL teachers for distillation, trained with log-replay on semantic camera or bird's-eye-view rasters~\cite{Jaeger2025CORL}, or by self-play on vectorized inputs~\cite{Pufferdrive2025github}. We confirm that strong RL teachers can be trained from semantic camera rasters alone~\cite{Yin2026ARXIV}, but do not find evidence that this camera perspective makes the teacher easier to distill than a bird's-eye-view teacher in our setting.
\textbf{(3)} An analysis of mixing with Sour, showing that human demonstrations partially recover the traffic-rule compliance and driving conventions that the teacher's inputs and reward do not cover. However, we find that domain gaps between simulator and real data can limit this effect for recorded camera logs.

\section{Related Work}
\label{sec:related}

\boldparagraph{Reinforcement Learning for Driving} Early RL methods for driving were mostly trained in slow game-engine simulators~\cite{Dosovitskiy2017CORL, Liang2018ECCVb, Toromanoff2020CVPR}, which hindered scaling. Lightweight simulators instead reduce the driving scenes to bounding boxes and HD maps, which enables scaling to billions of interactions~\cite{Cusumano2025ICML, Kazemkhani2025ICLR, Pufferdrive2025github, Jaeger2025CORL}. For instance, Gigaflow~\cite{Cusumano2025ICML} scales Proximal Policy Optimization (PPO)~\cite{Schulman2017ARXIV} with self-play in synthetic scenes, and CaRL~\cite{Jaeger2025CORL} scales log-replay simulation, both yielding state-of-the-art planners. SPICED~\cite{Cornelisse2026CORL} and related methods~\cite{Cornelisse2024RLJ} regularize RL with human demonstrations to keep human driving conventions. Recently, Pictura~\cite{Yin2026ARXIV} trained driving agents on perspective renderings of the abstract environment, so that the agents act only on what a camera could observe.
All these policies see the scene only through the abstraction and are rewarded on the same annotations. RL on camera inputs avoids the abstraction, but the number of interactions is limited by the cost of sensor simulation with 3DGS reconstructions or with world models~\cite{Gao2025NEURIPS, Yang2025NEURIPS, Ni2026ECCV, Li2026ARXIV}. 
In this work, we train a teacher with RL based on CaRL that, like Pictura, observes semantic rasters from the viewpoint of the student's cameras. We unroll the teacher to provide distillation targets for end-to-end driving models.
We mix human demonstrations into the distillation rather than into the teacher’s training, since the teacher observes the annotated scene and cannot relate human behavior to cues outside the annotation.

\boldparagraph{Distillation for End-to-End Driving} End-to-end policies are mostly trained by imitating real-world recordings of human demonstrators~\cite{Bojarski2016ARXIV, Hu2023CVPR, Kirby2026CVPR} or by imitating privileged experts in simulation~\cite{Chen2019CORL,Prakash2021CVPR,Zhang2021ICCV}. LEAD~\cite{Nguyen2026CVPR} shows that such experts act on information the student cannot perceive, and reduces this asymmetry by aligning the expert with the student's sensors. RAP~\cite{Feng2026ICLR} augments human logs with rasterized views of perturbed states, while Gigapixel~\cite{Rowe2026CORL}, DriveZero~\cite{He2026ARXIV} and TerraTransfer~\cite{Xiong2026ARXIV} replace human targets with RL teachers on vectorized inputs. These students train either in closed loop on simplified renderings or in open loop on recorded images. RoaD~\cite{Garcia2026CVPR} trains on closed-loop rollouts of the student that are guided towards the human log, which keeps the supervision close to the recorded drive. Concurrent to our work, OPTED~\cite{Da2026ARXIV} unrolls an RL teacher with vectorized inputs to label closed-loop rollouts in AlpaSim~\cite{Nvidia2025ALPASIM}, but shows improvements on weaker student policies and uses human demonstrations only for pre-training. We find that students distilled from an RL teacher inherit limitations of the teacher, and mix human demonstrations into the distillation to mitigate them.
\section{Background}
\label{sec:background}

We post-train a pre-trained end-to-end driving policy $\pi_\theta$, referred to as the student. At each step $t$, the student observes $o_t = O(s_t)$, the camera images and ego status of the scene state $s_t$, and predicts a trajectory $\tau_t = (w_1, \dots, w_H)$ of future waypoints~\cite{Chen2019CORL,Chitta2023PAMI, Hu2023CVPR, Wang2025ARXIV}, which a controller converts into an action $a_t$. In closed-loop driving, $\pi_\theta$ induces a distribution $d^{\pi_\theta}$ over the visited states $s_t$. Training objectives differ in the states on which the policy is trained and in the signal that supervises the policy at these states.

\boldparagraph{Behavior Cloning (BC)} The student imitates a demonstrator $\pi_E$, usually a human driver, by minimizing a training loss $\ell$ on the demonstrator's states,
\begin{equation}
\mathcal{L}_{\mathrm{BC}}(\theta) = \mathbb{E}_{s \sim d^{\pi_E}}\big[\ell\big(\pi_\theta(O(s)),\, \tau_E(s)\big)\big],
\label{eq:bc}
\end{equation}
where $\tau_E(s)$ is the target trajectory of the demonstrator at state $s$. The loss $\ell$ depends on the student, e.g., a distance for regression or a flow-matching objective with $\tau_E(s)$ as the clean sample for generative heads~\cite{Lipman2023ICLR}, as in our students~\cite{Li2026CORL, Nguyen2026CVPR}. The target reflects human driving, and training does not require a simulator. The supervision of BC is limited to $d^{\pi_E}$, whereas the student is deployed on $d^{\pi_\theta}$, leading to compounding errors as the student moves further from the training distribution~\cite{Ross2010AISTATS, Codevilla2019ICCV}.

\boldparagraph{Reinforcement Learning (RL)} Driving in RL is formulated as a Markov decision process $\mathcal{M} = (\mathcal{S}, \mathcal{A}, P, r)$, in which a simulator provides the transitions $P(s_{t+1} | s_t, a_t)$ and a reward $r(s_t, a_t)$ scores each step~\cite{Sutton1998MIT, Jaeger2024FTO}. A policy $\pi(a_t | s_t)$ maximizes the return $J(\pi) = \mathbb{E}_{\pi, P}\big[\sum_t \gamma^t r(s_t, a_t)\big]$ with discount factor $\gamma$. In contrast to BC, training and deployment share the state distribution $d^{\pi}$, resulting in closed-loop robustness and the ability to recover from mistakes. The supervision, however, is a scalar reward instead of a target trajectory. A hand-designed reward has to specify good driving~\cite{Knox2023Elsevier}, and exploration is needed to learn strong policies. Lightweight simulators of maps and boxes enable exploration at scale and yield robust privileged planners~\cite{Jaeger2025CORL,Cusumano2025ICML}. For camera-based policies, every transition requires rendering $O(s_{t+1})$ and running a large network~\cite{Wang2025ARXIV,Yang2025NEURIPS}, which makes exploration at this scale costly.

\section{Method}
\label{sec:method}

Our method combines both approaches from~\secref{sec:background}. We train an RL teacher $\pi^\star$ on privileged annotations, \ie, semantic camera or bird’s-eye-view (BEV) rasters (\secref{sec:method_rlteacher}). Pisco trains the student on the student's own closed-loop states $s \sim d^{\pi_\theta}$, as in RL, with the teacher's trajectories $\tau^\star(s)$ as targets, as in BC (\secref{sec:method_distillation}). Sour keeps the BC loss of Eq.~\eqref{eq:bc} on human demonstrations in every update.

\subsection{Reinforcement Learning Teacher}
\label{sec:method_rlteacher}

Pisco and Pisco Sour require a privileged teacher $\pi^\star$ that outputs targets for the student during on-policy rollouts. Privileged observations in BEV or vectorized states can include regions outside the student cameras' field-of-view, which creates a learner-expert asymmetry for imitation learning, as studied in LEAD~\cite{Nguyen2026CVPR}. We train our teacher $\pi^\star$ with PPO~\cite{Schulman2017ARXIV} and compare different observation spaces, including semantic camera rasters from the student’s viewpoint. We simulate with a kinematic bicycle model for the ego vehicle and non-reactive replay of other agents and objects in the scene. The teacher trains for 1B environment steps on 100k scenes from the nuPlan training split~\cite{Karnchanachari2024ICRA} and 16{,}070 NuRec scenes~\cite{Nvidia2026NUREC} that do not overlap with the scenes used for student distillation. 

\begin{figure}
    \centering
    \includegraphics[width=1.0\linewidth]{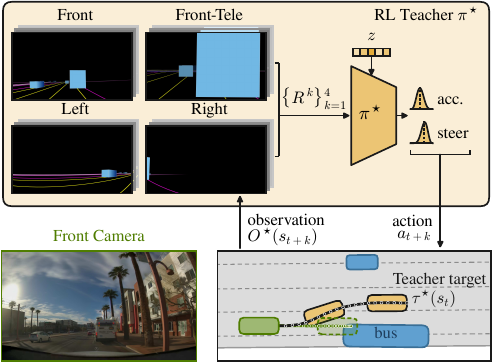}
    \vspace{-0.6cm}
    \caption{\textbf{Teacher Rollout.} The teacher $\pi^\star$ maps rasters $\{R^k\}$ of four cameras and measurements $z$ to acceleration and steering. We generate a target trajectory $\tau^\star(s_t)$, at a student state $s_t$, by rolling out $\pi^\star$ for 20 steps in co-simulation.}
    \label{fig:teacher_unroll}
    \vspace{-0.5cm}
\end{figure}

\boldparagraph{Observation} The teacher $\pi^\star(a | o^\star)$ observes $o^\star = O^\star(s) = (\{R^k\}_{k=1}^{4}, z)$: rasters $R^k$ from the four cameras of the PAI-AV rig~\cite{Nvidia2025PAIAV} (see \figref{fig:teacher_unroll}), and a measurement vector $z$ with the ego kinematics, the previous action, and route coordinates. Each raster stacks the last three frames and shows lane centerlines, road edges, crosswalks, stop zones, and surrounding agents as 3D boxes, as well as traffic lights in the nuPlan scenes, where the logs record the signal states.

\boldparagraph{Architecture} We embed the rasters $R^k$ with a shared convolutional encoder as in~\cite{Zhang2021ICCV, Jaeger2025CORL}. We pass all raster embeddings through a single cross-attention layer~\cite{Yin2026ARXIV, Vaswani2017NIPS}, together with a learned query embedding to pool all images. Following~\cite{Jaeger2025CORL}, an MLP fuses the pooled embedding with $z$ and outputs Beta distributions over normalized acceleration and steering angle, which the ego bicycle model executes at 10\,Hz. Our teacher has 6.2M parameters.

\boldparagraph{Reward} Our reward uses a combination of multiplier penalties and a weighted average of driving metrics:
\begin{equation}
r(s_t, a_t) = \prod_i m_i(s_t) \cdot \sum_j \lambda_j\, c_j(s_t, a_t).
\label{eq:reward}
\end{equation}
The metrics $c_j \in [0, 1]$ are route progress, time-to-collision, speed-limit compliance, and comfort, with weights $\lambda_j$ from the nuPlan closed-loop score~\cite{Karnchanachari2024ICRA}, normalized to $\sum_j \lambda_j = 1$. The penalties $m_i \in [0, 1]$ reduce the reward for leaving the route, deviating from the lane center, and falling behind the progress of the log. At-fault collisions and leaving the road end the episode with zero reward. The reward has no traffic-light term, as our training scenes inconsistently provide annotations for their states. The supplementary material details the observation, architecture, reward, and training hyperparameters.

\subsection{On-Policy Distillation}
\label{sec:method_distillation}

On-policy distillation commonly matches the output distributions of teacher and student by minimizing a Kullback--Leibler divergence~\cite{Agarwal2024ICLR, Gu2024ICLR, Rusu2016ICLR}. Our RL teacher outputs actions, as is common for RL driving policies~\cite{Jaeger2025CORL, Cusumano2025ICML, Zhang2021ICCV}, whereas end-to-end driving models typically output trajectories~\cite{Chitta2023PAMI,Hu2023CVPR,Wang2025ARXIV,Kirby2026CVPR,Dauner2024NEURIPS}. We therefore unroll the teacher instead (\figref{fig:teacher_unroll}). We copy the visited state $s$ into the teacher's simulator and unroll $\pi^\star$ for $H^\star = 20$ steps (2\,s) with the mean action, where the other agents are replayed from the log. The ego poses of the rollout form the target $\tau^\star(s) = (w^\star_1, \dots, w^\star_{H^\star})$. We then minimize
\begin{equation}
\mathcal{L}_{\mathrm{Pisco}}(\theta) = \mathbb{E}_{s \sim d^{\pi_\theta}}\big[\ell\big(\pi_\theta(O(s))_{1:H^\star},\, \tau^\star(s)\big)\big],
\label{eq:pisco}
\end{equation}
where the subscript $1{:}H^\star$ selects the first $H^\star$ waypoints of the student's plan $\tau = (w_1, \dots, w_H)$. For a flow-matching head, the clean sample is completed after $H^\star$, and the loss is masked to the first $H^\star$ waypoints, so later waypoints receive no teacher supervision.

\boldparagraph{Training Loop} We outline our training in Algorithm~\ref{alg:pisco}. The student drives $K$ scenes in parallel in neural reconstructions of recorded drives~\cite{Nvidia2026NUREC,Li2025ARXIV}, and the teacher is rolled out at every state visited by the student (\figref{fig:teacher_unroll}). 
After these rollouts, Sour adds human pairs from a dataset $\mathcal{D}$ to make up a share $f$ of all pairs (\eg, half of the pairs at $f = 0.5$, doubling the optimizer steps).
The student takes one pass over all pairs and minimizes
\begin{equation}
\mathcal{L}_{\mathrm{Pisco\,Sour}}(\theta) = (1-f)\,\mathcal{L}_{\mathrm{Pisco}}(\theta) + f\,\mathcal{L}_{\mathrm{BC}}(\theta).
\label{eq:sour}
\end{equation}
In $\mathcal{L}_{\mathrm{BC}}$, the human target supervises all $H$ waypoints. The samples are then discarded, so the on-policy pairs of every update come from the current student. The updated student drives the next $K$ scenes, for $N$ passes over the complete training scene set $\mathcal{X}$.

\boldparagraph{Data Aggregation} As a variant, Pisco$_{+}$ aggregates the teacher-labeled data over multiple rounds, as in DAgger~\cite{Ross2011AISTATS}. In each of $N$ rounds, the current student drives all training scenes in $\mathcal{X}$, the teacher labels every visited state, and the student trains for one pass over the data of all rounds so far. Pisco$_{+}$ Sour initializes the aggregated dataset with the human demonstrations of all training scenes. Over $N$ rounds, Pisco$_{+}$ Sour therefore uses the same student environment interactions and trains $N$ times on every human demonstration, like Pisco Sour at $f=0.5$.

\begin{algorithm}[t]
\caption{Pisco and \sour{Sour}.}
\label{alg:pisco}
\algrenewcommand\algorithmicindent{1.0em}
\small
\begin{algorithmic}[1]
\Require student $\pi_\theta$, teacher $\pi^\star$, scenes $\mathcal{X}$, passes $N$, parallel scenes $K$\sour{, human data $\mathcal{D}$, share $f$}
\For{$n = 1, \dots, N$} \Comment{passes over $\mathcal{X}$}
    \For{each batch of $K$ scenes of shuffled $\mathcal{X}$}
        \State $\mathcal{B} \gets \emptyset$
        \For{each scene of the batch, in parallel}
            \State $s \gets$ initial state of the scene
            \While{scene not finished}
                \State $o \gets O(s)$ \Comment{neural rendering}
                \State $\tau^\star \gets \textsc{Unroll}(\pi^\star, s, H^\star)$ \Comment{teacher, \figref{fig:teacher_unroll}}
                \State $\mathcal{B} \gets \mathcal{B} \cup \{(o, \tau^\star)\}$
                \State $a \gets \textsc{Control}(\pi_\theta(o))$ \Comment{student}
                \State $s \gets P(s, a)$
            \EndWhile
        \EndFor
        \State \sour{$\mathcal{B} \gets \mathcal{B} \cup \{\tfrac{f}{1-f}|\mathcal{B}| \text{ pairs } (o, \tau_E) \sim \mathcal{D}\}$}
        \For{each minibatch $b \subset \mathcal{B}$}
            \State update $\theta$ on $\sum_{(o, \tau) \in b} \ell(\pi_\theta(o), \tau)$
        \EndFor
    \EndFor
\EndFor
\end{algorithmic}

\end{algorithm}
\section{Experiments}
\label{sec:experiments}

We first describe our experimental setup and compare Pisco Sour with other post-training strategies in closed-loop evaluation (\secref{sec:closed_loop}). We further compare different teachers (\secref{sec:teacher_comparison}). Next, we study which behaviors the student inherits from the teacher and how human drives counteract them (\secref{sec:human_mixing}). Finally, we evaluate the students on other closed-loop benchmarks (\secref{sec:cross_benchmark}).

\subsection{Closed-Loop Training}
\label{sec:closed_loop}

\boldparagraph{Students}
We post-train two students. (1) TransFuser~\cite{Nguyen2026CVPR} with three input cameras (front wide, cross left and right) with 68M parameters, pre-trained on PAI-AV~\cite{Nvidia2025PAIAV} and nuPlan~\cite{Karnchanachari2024ICRA} recordings, following the training recipe of \texttt{py123d\_garage}~\cite{Dauner2026ARXIV}. TransFuser receives the current and 1\,s old frames, together with route target points 40, 60 and 80\,m ahead, and predicts a 6.4\,s trajectory with a flow-matching head. We train the whole network by default. (2) FlashDrive~\cite{Li2026CORL} is an optimized variant of the 10B vision-language-action model Alpamayo~1.5~\cite{Wang2025ARXIV}, with the VLM quantized to 4-bit weights and 8-bit activations for faster inference~\cite{Liang2026ICLR}. FlashDrive reads four frames of four PAI-AV cameras (front wide, front tele, cross left and right) and drives without a route condition. We post-train only the 2.3B flow-matching head and disable the reasoning trace for faster inference. Implementation details are provided in the supplementary material. 

\boldparagraph{Baselines}
We compare against two post-training strategies. (1) Supervised fine-tuning (SFT) trains on the re-rendered human drives of the training scenes for 4 epochs, to measure the real-to-sim gap of each student. (2) RoaD~\cite{Garcia2026CVPR} samples 64 trajectories from the student at each step, executes the one closest to the human log over the first 2\,s (with recovery trajectories at $>$3\,m deviation), and trains on the executed 6.4\,s trajectories. We post-train each student with both variants of our method from~\secref{sec:method_distillation}, Pisco and Pisco$_{+}$, each with and without Sour. 
All closed-loop methods train for $N=4$ passes over the training scenes, \ie, 1.4M environment steps vs.\ 1B for the teacher.
We further evaluate public model releases of VaVAM~\cite{Bartoccioni2025ARXIV}, Qwen-Drive~1.0~\cite{Zhou2026ARXIV} and Alpamayo~2~Super~\cite{Nvidia2026ALPAMAYO2SUPER} as references on the validation set.

\boldparagraph{Benchmark \& Metrics}
We evaluate in AlpaSim~\cite{Nvidia2025ALPASIM} on the official Val441 set, which includes $441$ NuRec reconstructions~\cite{Nvidia2026NUREC} of 20\,s duration each. 
All methods post-train on the 1{,}761 NuRec training scenes, which share no clips with Val441.
AlpaSim assigns each scene $v$ a score $\sigma_v$. A scene fails on an at-fault collision, an off-road event, or a lateral corridor deviation of more than 4\,m from the recorded path. Since the other agents are non-reactive, only front and lateral collisions count as at fault. With $\rho_v$ the progress along the recorded path relative to the log, the scene score is
\begin{equation}
\sigma_v =
\begin{cases}
0 & \text{if scene } v \text{ fails},\\
\min\!\big(\rho_v / 0.8,\, 1\big) & \text{otherwise},
\end{cases} \in [0,1]
\label{eq:score}
\end{equation}
so progress counts fully from 80\,\% on. We further report the distance driven per at-fault collision or off-road event (km/inf.), which normalizes these failures by the distance driven, the share of scenes with each failure type (at-fault collisions, off-road, corridor), and the mean progress.

\boldparagraph{Results} In~\tabref{tab:alpasim_main}, we compare both students post-trained with Pisco with and without aggregation or adding Sour, against the baselines and reference models. Both pre-trained students already outperform the strongest reference model, Alpamayo~2 Super (59.3 and 69.9 vs.\ 53.0\,\% score). Our two base students have different failure modes, \eg, TransFuser goes off-road less than FlashDrive (5.0 vs.\ 13.4\,\%). Models such as TransFuser with target point inputs often learn to bias their trajectories towards the navigation coordinates~\cite{Jaeger2023ICCV}, with the side effect of lower off-road rates. Without target points, the base TransFuser leaves the road and the route corridor more often (see supplementary material). The base TransFuser, however, has low progress in comparison (70.1 vs.\ 80.4\,\%). SFT mostly raises the progress of TransFuser (70.1 to 83.0\,\%), while the off-road rate stays similar.
Since the other actors replay the log, a standing ego vehicle can be hit from the front or side, which counts as at fault. SFT keeps pace with the log progress and avoids most of these collisions (4.5 to 2.7\,\%, see supplementary material).
In contrast, FlashDrive shows less improvement from SFT with lower collision rates but an increase in off-road events. Since closed-loop strategies also train on rendered data, adapting to the simulator is part of why such methods can improve in AlpaSim. However, SFT does not resolve all failure modes (\eg, collisions), and FlashDrive barely benefits from SFT at all.

\begin{table}[t]
    \centering
    \footnotesize
    \setlength{\tabcolsep}{1pt}
    \newcommand*{\hd}[2]{\multicolumn{1}{c}{\begin{tabular}[b]{@{}c@{}}#1\\#2\end{tabular}}}
    \begin{tabular*}{\linewidth}{@{\extracolsep{\fill}}lrrrrrrrr@{}}
        \toprule
        & \multicolumn{6}{c}{Closed-loop} & \multicolumn{2}{c}{ADE$_{2\text{s}}$} \\
        \cmidrule(lr){2-7} \cmidrule(l){8-9}
        Method & \hd{Score}{$\uparrow$} & \hd{km/inf.}{$\uparrow$} & \hd{Coll.}{$\downarrow$} & \hd{Offr.}{$\downarrow$} & \hd{Corr.}{$\downarrow$} & \hd{Prog.}{$\uparrow$} & \hd{NuRec}{$\downarrow$} & \hd{Real}{$\downarrow$} \\
        \midrule
           VaVAM~\cite{Bartoccioni2025ARXIV}               & 6.9 & 0.08 & 17.2 & 37.0 & 44.0 & 32.1 & -- & -- \\
        Qwen-Drive 1.0~\cite{Zhou2026ARXIV}       & 22.1 & 0.19 & 15.9 & 22.2 & 21.3 & 39.0 & -- & -- \\
        Alpa.~2 Super~\cite{Nvidia2026ALPAMAYO2SUPER}       & 53.0 & 0.84 & 5.2 & 12.2 & 26.5 & 73.8 & -- & -- \\
        \midrule
        TransFuser~\cite{Nguyen2026CVPR}                        & 69.9 & 1.70 & 4.5 & 5.0 & \textbf{1.8} & 70.1 & 0.34 & \textbf{0.26} \\
        \quad SFT                        & 82.1 & 2.40 & 2.7 & 4.8 & 3.6 & 83.0 & \textbf{0.28} & \textbf{0.26} \\
        \quad RoaD~\cite{Garcia2026CVPR}                       & 77.4 & 1.70 & 4.1 & 5.9 & 2.0 & 77.8 & 0.31 & \textbf{0.26} \\
        \quad Pisco      & 83.8 & 2.08 & 4.5 & 3.9 & 3.6 & 85.0 & 0.59 & 0.56 \\
        \quad Pisco$_{+}$      & 85.7 & 2.71 & \textbf{2.5} & 4.1 & 4.1 & 86.7 & 0.44 & 0.42 \\
        \quad Pisco Sour & \textbf{86.9} & \textbf{3.11} & 3.2 & 2.7 & 3.2 & \textbf{87.7} & 0.35 & 0.34 \\
        \quad Pisco$_{+}$ Sour & 86.3 & 2.86 & 3.9 & \textbf{2.5} & 3.9 & 87.5 & 0.35 & 0.34 \\
        \midrule
        FlashDrive~\cite{Li2026CORL}                        & 59.3 & 0.88 & 5.4 & 13.4 & 20.0 & 80.4 & 0.22 & 0.18 \\
        \quad SFT                        & 60.4 & 0.77 & 2.5 & 16.8 & 15.6 & 75.1 & \textbf{0.16} & \textbf{0.15} \\
        \quad RoaD~\cite{Garcia2026CVPR}                        & 68.7 & 1.41 & 3.2 & 9.3 & 15.6 & 82.9 & 0.20 & 0.18 \\
        \quad Pisco      & 79.5 & 4.05 & 2.3 & 2.3 & 13.6 & 86.2 & 0.41 & 0.36 \\
        \quad Pisco$_{+}$      & 81.6 & \textbf{5.95} & 1.4 & \textbf{1.8} & 12.9 & 87.7 & 0.39 & 0.35 \\
        \quad Pisco Sour & \textbf{83.5} & 5.58 & \textbf{0.7} & 2.7 & \textbf{11.6} & 88.8 & 0.23 & 0.21 \\
        \quad Pisco$_{+}$ Sour & 81.9 & 4.32 & 2.0 & 2.5 & 12.2 & \textbf{89.3} & 0.25 & 0.21 \\
        \bottomrule
    \end{tabular*}
    \vspace{-0.1cm}
    \caption{\textbf{Closed-loop results in AlpaSim} (Val441). The FlashDrive and TransFuser rows show the base models. Pisco: on-policy distillation, Pisco$_{+}$: with data aggregation as in DAgger~\cite{Ross2011AISTATS}. Score,  at-fault collisions (Coll.), off-road (Offr.), corridor exits (Corr.) and progress (Prog.) in \%; ADE$_{2\text{s}}$ to the human drive in m, on NuRec renders and recorded PAI-AV frames.}
    \label{tab:alpasim_main}
    \vspace{-0.5cm}
\end{table}

\begin{figure*}
    \centering
    \includegraphics[width=1.0\linewidth]{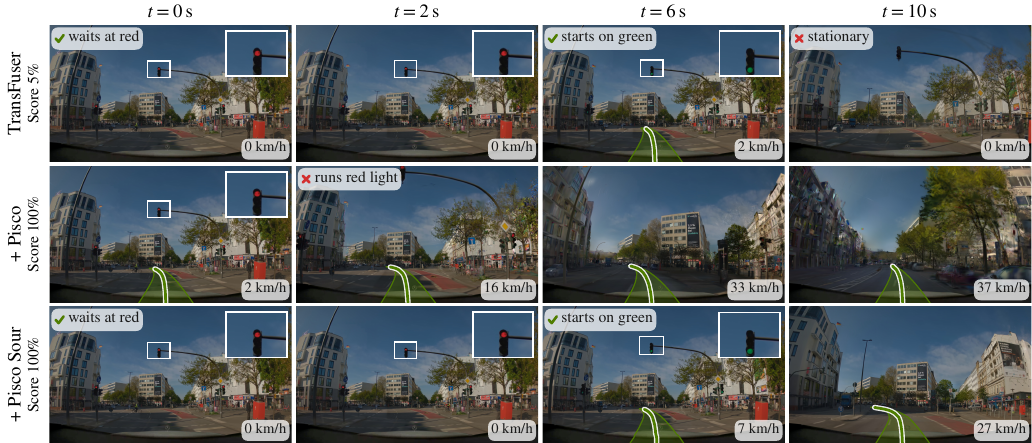}
    \caption{\textbf{Pisco vs.\ Pisco Sour.}
    Closed-loop TransFuser rollouts in AlpaSim (plan in green with car-width polygon).
    The base TransFuser waits at the red light but barely moves on green and becomes stationary (score 5\%).
    The teacher of Pisco nudges the student towards progress but cannot perceive traffic lights, so the distilled policy runs the red light and still scores 100\%, since the AlpaSim score does not consider traffic lights.
    With human data mixed in, Pisco Sour waits at the red light and starts on green, also receiving a 100\% score.}
    \label{fig:qualitative}
    \vspace{-0.5cm}
\end{figure*}

Closed-loop training with RoaD improves the score of FlashDrive from 59.3 to 68.7\,\% and reduces off-road events from 13.4 to 9.3\,\%. This indicates that training on the student's own state distribution can improve recovery behavior. For TransFuser, RoaD raises the score from 69.9 to 77.4\,\%, below SFT. With target points as input, we find that the trajectory distribution of TransFuser is less diverse, which limits RoaD in selecting suitable candidate target trajectories.

In contrast to SFT and RoaD, supervision from our teacher improves both students considerably. The effect is largest for FlashDrive, where Pisco raises the score from 59.3 to 79.5\,\% and decreases off-road events (13.4 to 2.3\,\%). For TransFuser, the improvement stems mainly from higher progress (70.1 to 85.0\,\%), whereas the collision rate remains similar to the base model (\ie, 4.5\,\%). Aggregating the data of all previous rounds improves both students further: Pisco$_{+}$ reaches scores of 81.6 and 85.7\,\% and increases the distance per infraction to 5.95 and 2.71\,km. As Pisco$_{+}$ trains largely on states visited in earlier iterations, the higher scores suggest that restricting supervision to the current policy is not advantageous in our setting. However, Pisco$_{+}$ processes 2.5$\times$ the gradient samples of Pisco, which makes training more expensive.

We obtain the strongest students by adding human supervision with Sour. Both students progress further and collide less than with Pisco alone. Part of this gain could come from more updates, as Sour processes about 2$\times$ the gradient samples of Pisco at $f = 0.5$. Pisco$_{+}$, however, trains on even more samples (2.5$\times$) and does not exceed the score of Pisco Sour for either student. Adding Sour to Pisco$_{+}$ raises the progress but barely the score, and Pisco$_{+}$ Sour does not improve over Pisco Sour.

Teacher and human supervision have complementary effects, which we analyze further in~\secref{sec:human_mixing}. The open-loop ADE indicates that the students keep the human behavior with Sour. Pisco moves the plans of both students away from the human drive (0.22 to 0.41\,m for FlashDrive and 0.34 to 0.59\,m for TransFuser). With Sour, the plans stay as close to the human as those of the pre-trained students, with and without aggregation, and with the same trend on the real PAI-AV frames. With more updates but without human data, the plans of Pisco$_{+}$ remain far from the human drive (0.39 and 0.44\,m).

\begin{table}[t]
    \centering
    \small
    \setlength{\tabcolsep}{3pt}
    \resizebox{\linewidth}{!}{%
    \begin{tabular}{lrrrrrr}
        \toprule
        Method & Score$\uparrow$ & km/inf.$\uparrow$ & Coll.$\downarrow$ & Offr.$\downarrow$ & Corr.$\downarrow$ & Prog.$\uparrow$ \\
        \midrule
        TransFuser                           & 69.9 & 1.70 & 4.5 & 5.0 & 1.8 & 70.1 \\
        SFT                              & 82.1 & 2.40 & 2.7 & 4.8 & 3.6 & 83.0 \\
        \midrule
        \textit{PufferDrive}~\cite{Pufferdrive2025github}     & \textit{87.7} & \textit{6.52} & \textit{0.9} & \textit{1.8} & \textit{4.3} & \textit{85.7} \\
        \quad Pisco       & 83.4 & 1.88 & 3.4 & 5.7 & \textbf{2.7} & 84.4 \\
        \quad Pisco$_{+}$       & 83.6 & 2.30 & \textbf{3.2} & 4.5 & 4.5 & 85.3 \\
        \quad Pisco Sour  & 83.4 & 2.11 & 4.1 & 4.5 & 4.5 & 86.7 \\
        \quad Pisco$_{+}$ Sour  & \textbf{83.8} & \textbf{2.36} & 3.4 & \textbf{4.3} & 5.0 & \textbf{86.9} \\
        \midrule
        \textit{CaRL}~\cite{Jaeger2025CORL}            & \textit{87.1} & \textit{3.44} & \textit{4.1} & \textit{1.4} & \textit{3.9} & \textit{88.8} \\
        \quad Pisco       & 82.3 & 2.04 & 2.9 & 5.2 & 2.9 & 83.2 \\
        \quad Pisco$_{+}$       & 84.8 & 2.33 & 3.6 & 3.9 & 3.2 & 85.7 \\
        \quad Pisco Sour  & 86.0 & 2.50 & 2.9 & 4.3 & \textbf{2.5} & 86.6 \\
        \quad Pisco$_{+}$ Sour  & \textbf{86.7} & \textbf{3.20} & \textbf{2.3} & \textbf{3.4} & 3.4 & \textbf{87.2} \\
        \midrule
        \textit{Sem. Camera (ours)}      & \textit{89.9} & \textit{6.50} & \textit{1.4} & \textit{1.6} & \textit{3.9} & \textit{90.2} \\
        \quad Pisco       & 83.8 & 2.08 & 4.5 & 3.9 & 3.6 & 85.0 \\
        \quad Pisco$_{+}$       & 85.7 & 2.71 & \textbf{2.5} & 4.1 & 4.1 & 86.7 \\
        \quad Pisco Sour  & \textbf{86.9} & \textbf{3.11} & 3.2 & 2.7 & \textbf{3.2} & \textbf{87.7} \\
        \quad Pisco$_{+}$ Sour  & 86.3 & 2.86 & 3.9 & \textbf{2.5} & 3.9 & 87.5 \\
        \bottomrule
    \end{tabular}
    }
    \vspace{-0.1cm}
    \caption{
        \textbf{Choice of teacher.} Each privileged teacher drives by a \mbox{2-second} trajectory unroll;
        the indented rows below it are TransFuser post-trained from that teacher.
        Columns as in Table~\ref{tab:alpasim_main}.
    }
    \label{tab:alpasim_teachers}
    \vspace{-0.8cm}
\end{table}

\begin{figure*}[t]
    \centering
    \includegraphics[width=1.0\linewidth]{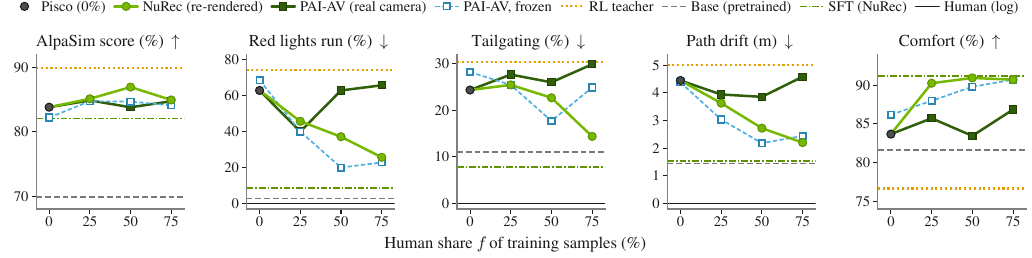}
    \vspace{-0.65cm}
    \caption{\textbf{Human mixing.} Closed-loop metrics of TransFuser over the human share $f$ of training samples, for re-rendered NuRec drives and recorded PAI-AV camera logs with a trainable or frozen backbone. At $f=0$, the student trains on teacher targets only (Pisco).}
    \label{fig:human_safety}
    \vspace{-0.325cm}
\end{figure*}

\subsection{Teacher Comparison}
\label{sec:teacher_comparison}

We compare our semantic camera teacher from~\secref{sec:method_rlteacher} with two other teachers. (1) PufferDrive~\cite{Pufferdrive2025github} is trained by self-play in procedurally generated traffic on 128 transformed variants of the eight CARLA towns~\cite{Dosovitskiy2017CORL}, for 15.6B agent steps. PufferDrive observes vectorized lanes, road boundaries, other agents and traffic controls, and drives towards goal points along the route. (2) CaRL~\cite{Jaeger2025CORL} takes a 256$\times$256 bird's-eye-view raster as input, at a resolution of 0.5\,m per pixel. We train CaRL with the same reward, scenes, and number of environment steps (1B) as our teacher. For each teacher, we post-train TransFuser with Pisco and Pisco$_{+}$, each with and without Sour, as in~\secref{sec:closed_loop}. We evaluate the teachers in the same closed-loop setting as the students. As for the distillation targets, each teacher drives by unrolling a 2\,s trajectory in co-simulation, which the controller then follows. During the unroll, the other agents are replayed from the log, so the teacher sees their future motion.

\boldparagraph{Results}
All privileged teachers have higher performance than the students in~\tabref{tab:alpasim_teachers}. Our semantic camera teacher achieves the highest score (89.9\,\%) and progress (90.2\,\%). PufferDrive, trained only with self-play, reaches 87.7\,\% and collides least (0.9\,\%), but progresses less (85.7\,\%), as PufferDrive often stops behind parked vehicles. The raster of CaRL is coarse and reaches only 78\,m ahead, leading to a higher collision rate (4.1\,\% vs.\ 1.4\,\% for our teacher), especially at high speeds and in tight maneuvers.

The TransFuser students reach 82.3 to 83.8\,\% with Pisco and 83.6 to 85.7\,\% with Pisco$_{+}$. With Pisco alone, the students score only slightly above SFT (82.1\,\%), which also adapts TransFuser to the rendered scenes. Adding Sour raises the score again for the semantic camera teacher and CaRL (to 86.9 and 86.0\,\%), whereas the PufferDrive student remains at 83.4\,\%. Interestingly, all students of CaRL collide less than the teacher itself (2.3 to 3.6 vs.\ 4.1\,\%). The students keep part of the larger following distances learned in pre-training and avoid most of CaRL’s collisions.

While our semantic camera teacher performs best in three of the four distillation settings, we do not find that the camera perspective makes the teacher easier to distill. The gap between our teacher and student remains substantial (89.9 vs.\ 83.8\,\%). As all three teachers create targets with knowledge of the future motion of other agents, there remains a fundamental asymmetry between the student's and teacher's inputs, even when their perspective is aligned. %

\subsection{Blending Human Demonstrations}
\label{sec:human_mixing}

Our teacher has limitations, as both the observation and the reward function are computed from scene annotations. These annotations can be incomplete, \eg, NuRec does not provide traffic-light states, and a reward on these annotations has to weigh all intended behaviors against each other, \eg, safety against progress and comfort. We next study how blending in human demonstrations can prevent the student from imitating undesired behavior of the teacher.

We post-train TransFuser with Pisco Sour at human shares $f \in \{0.25, 0.5, 0.75\}$, where $f=0$ corresponds to Pisco. We compare two sources of human drives: NuRec re-renderings of the human drives in the 1{,}761 training scenes, which match the appearance of the simulation, and the recorded camera frames of 150k PAI-AV training clips~\cite{Nvidia2025PAIAV} used during pre-training. For PAI-AV mixing, we additionally freeze the backbone and train only the planning decoder (15.5M of the 68M parameters), which limits the model's capacity to distinguish rendered from real data and overfit to specific behaviors for each. All other experiments use $f=0.5$, which we fixed before this study. 

\boldparagraph{Metrics} We use the recorded human drive of each scene as the reference to infer traffic-rule compliance and the following metrics.
\emph{Red lights:} If the log vehicle waits at a traffic-light stop line without a vehicle in front, we treat the light as red (35 scenes). The policy runs the light if the policy crosses the line while the log vehicle brakes for or waits at the line. We provide thresholds and visualizations of the events in the supplementary material.
\emph{Tailgating:} The policy tailgates if the policy keeps a time headway (in seconds) to the lead vehicle below 0.7$\times$ that of the log vehicle (181 scenes).
\emph{Path drift:} Mean distance of the policy from the path of the log vehicle.
\emph{Comfort:} Share of comfortable scenes under the nuPlan comfort metric~\cite{Karnchanachari2024ICRA}, with thresholds for acceleration, jerk, yaw rate and yaw acceleration.

\boldparagraph{Results}
\figref{fig:human_safety} shows that students trained only with Pisco inherit the limitations of the teacher. At $f=0$, TransFuser runs 63\,\% of the red lights and tailgates in 24\,\% of the scenes, similar to the RL teacher (74\,\% and 30\,\%). The base and SFT models rarely run red lights (3 and 9\,\%) or tailgate (11 and 8\,\%). Adding Sour with re-rendered human demonstrations reduces these deviations as the human share grows. From $f=0$ to $f=0.75$, red-light violations fall from 63 to 26\,\%, tailgating from 24 to 14\,\%, and path drift from 4.4 to 2.2\,m. Sour also improves comfort from 84 to 90\,\% already at $f=0.25$. The AlpaSim score stays above Pisco and SFT, varying little with $f$ (84.9 to 86.9\,\%).

Sour with real camera logs helps only partially. With PAI-AV drives, red-light violations drop only at $f=0.25$ (40\,\%) and rise again to 66\,\% at $f=0.75$, while tailgating (28 to 30\,\%) and path drift (3.9 to 4.6\,m) show no improvement over Pisco. We expect that the student distinguishes real from rendered observations and imitates either the human demonstrator or the teacher accordingly. Freezing the backbone when mixing real PAI-AV behaves similarly to Sour on rendered NuRec data. The score, however, is lower (84.6 vs.\ 86.9\,\% at $f=0.5$), since the frozen backbone cannot adapt to the simulator and closed-loop states of the student. Moreover, prior research suggests that even frozen image encoders can identify the source dataset of an image~\cite{Liu2025ICLR}, so freezing cannot ensure that the student treats real and rendered images alike.
Human mixing is effective, as we show with rendered data, but Sour with real data remains challenging due to the visual sim-to-real gap.

\begin{table}[t]
    \centering
    \footnotesize
    \setlength{\tabcolsep}{2.5pt}
    \setlength{\aboverulesep}{0pt}
    \setlength{\belowrulesep}{0pt}
    \renewcommand{\arraystretch}{1.2}
    \begin{tabular*}{\linewidth}{@{\extracolsep{\fill}}l>{\columncolor{gray!15}[4pt][4pt]}rrrrrrr@{}}
        \toprule
        & \multicolumn{5}{c}{HD-Score\,$\uparrow$} & & \\
        \cmidrule(lr){2-6}
        Method & \textbf{Avg.} & E & M & H & X & RC\,$\uparrow$ & C\,$\uparrow$ \\
        \midrule
        VAD~\cite{Jiang2023ICCV}                   & 13.4 & 36.3 & 9.5 & 8.0 & 11.5 & 31.4 & \textcolor{gray}{--} \\
        LTF~\cite{Chitta2023PAMI}                  & 23.7 & 58.9 & 18.0 & 9.8 & 25.9 & 38.9 & \textcolor{gray}{--} \\
        UniAD~\cite{Hu2023CVPR}                    & 32.7 & 64.9 & 45.8 & 20.6 & 6.6 & 45.9 & \textcolor{gray}{--} \\
        DrivoR~\cite{Kirby2026CVPR}                & 35.7 & 73.3 & 34.6 & 18.8 & \textbf{32.5} & 49.8 & \textcolor{gray}{--} \\
        DrivoR-Scale~\cite{Kirby2026CVPR}          & 38.1 & 64.8 & 50.3 & 16.8 & 25.8 & 46.4 & \textcolor{gray}{--} \\
        Gigapixel~\cite{Rowe2026CORL}              & 38.5 & 67.4 & \underline{51.9} & 19.1 & 21.6 & 50.1 & \textcolor{gray}{--} \\
        DriveZero~\cite{He2026ARXIV}               & 39.4 & \underline{78.5} & 47.2 & 16.0 & 28.7 & 53.9 & \textcolor{gray}{--} \\
        DriveZero-Scale~\cite{He2026ARXIV}         & \textbf{46.6} & \textbf{81.9} & \textbf{60.5} & \underline{24.7} & 27.6 & \underline{54.5} & \textcolor{gray}{--} \\
        \midrule
        TransFuser~\cite{Nguyen2026CVPR}           & 30.8 & 52.1 & 46.5 & 19.3 & 6.6 & 37.4 & \textbf{99.9} \\
        \quad Pisco                                & \underline{42.5} & 76.1 & 50.8 & 24.3 & \underline{29.0} & \textbf{58.6} & 94.0 \\ %
        \quad Pisco Sour                           & 40.4 & 77.1 & 46.7 & \textbf{28.6} & 21.8 & 53.4 & \underline{97.6} \\ %
        \bottomrule
    \end{tabular*}
    \vspace{-0.1cm}
    \caption{\textbf{Closed-loop results on HUGSIM~\cite{Zhou2025PAMI}.} HD-Score on easy (E), medium (M), hard (H) and extreme (X) scenarios, route completion (RC) and comfort (C). Results with the same protocol are taken from~\cite{Kirby2026CVPR, Rowe2026CORL, He2026ARXIV}.}
    \label{tab:hugsim_internal}
    \vspace{-0.5cm}
\end{table}

\subsection{Cross-Benchmark Evaluation}
\label{sec:cross_benchmark}

We complement our AlpaSim evaluation with two further closed-loop benchmarks. \textbf{(1) HUGSIM}~\cite{Zhou2025PAMI} contains the 345 pre-challenge scenarios of nuScenes~\cite{Caesar2020CVPR}, PandaSet~\cite{Xiao2021ITSC}, KITTI-360~\cite{Liao2022PAMI} and Waymo~\cite{Sun2020CVPR} in four difficulty levels. We report the HD-Score, route completion (RC) and comfort (C). The HD-Score combines the route completion with averaged scores for no collisions, drivable area compliance, time to collision and comfort. \textbf{(2) Odyssey}~\cite{Kim2026ARXIV} contains 100 nuPlan test scenes in non-reactive and reactive traffic. The route driving score (RDS) combines the route completion with penalties for collisions, off-road driving and traffic violations. We further report the share of episodes that stay on the route (SDC) and the share of comfortable episodes (C). 
For this evaluation, we post-train TransFuser on the NuRec scenes and 12,134 additional nuPlan training scenes of 8 s duration, rendered with MTGS~\cite{Li2025ARXIV}, and Sour mixes in the recorded camera logs of the PAI-AV and nuPlan training splits, none of which overlap with Odyssey.

\boldparagraph{Results}
Pisco improves the base model on both benchmarks. On HUGSIM (\tabref{tab:hugsim_internal}), Pisco raises the HD-Score of TransFuser from 30.8 to 42.5 and Pisco Sour to 40.4, second only to DriveZero-Scale (46.6). 
On Odyssey (\tabref{tab:odyssey_internal}), Pisco and Pisco Sour reach the highest RDS of all methods (59.8 and 59.6 non-reactive, 58.2 and 64.1 reactive), above SafeDrive with test-time safety filtering (54.4 and 56.1). 
In reactive traffic of Odyssey, the gain of Pisco Sour over Pisco comes from fewer collisions and better route compliance.
On both benchmarks, Sour keeps the score of Pisco and drives more comfortably (94.0 to 97.6 on HUGSIM, 40 to 64\,\% comfortable episodes on Odyssey). 
Without training on additional nuPlan scenes, the gains are smaller, especially on Odyssey, whose simulations come from the nuPlan test split (see supplementary material).

\begin{table}[t]
    \centering
    \footnotesize
    \setlength{\tabcolsep}{2.5pt}
    \setlength{\aboverulesep}{0pt}
    \setlength{\belowrulesep}{0pt}
    \renewcommand{\arraystretch}{1.2}
    \begin{tabular*}{\linewidth}{@{\extracolsep{\fill}}l>{\columncolor{gray!15}[4pt][4pt]}rrr>{\columncolor{gray!15}[4pt][4pt]}rrr@{}}
        \toprule
        & \multicolumn{3}{c}{Non-reactive} & \multicolumn{3}{c}{Reactive} \\
        \cmidrule(lr){2-4}\cmidrule(l){5-7}
        Method & \textbf{RDS}\,$\uparrow$ & SDC\,$\uparrow$ & C\,$\uparrow$ & \textbf{RDS}\,$\uparrow$ & SDC\,$\uparrow$ & C\,$\uparrow$ \\
        \midrule
        LTF~\cite{Chitta2023PAMI}                  & 20.1 & 95 & 96 & 23.0 & 93 & 93 \\
        ReCogDrive~\cite{Xiong2026ICLR}            & 21.5 & 83 & 98 & 26.3 & 84 & 98 \\
        DiffusionDrive~\cite{Liao2025CVPR}         & 35.9 & \textbf{100} & \underline{99} & 41.0 & \textbf{100} & \underline{99} \\
        DrivoR~\cite{Kirby2026CVPR}                & 36.5 & 90 & 60 & 41.4 & 89 & 64 \\
        SafeDrive~\cite{Kim2026CVPR}               & 44.4 & 82 & 44 & 48.9 & 79 & 49 \\
        SafeDrive + filter~\cite{Kim2026CVPR}      & 54.4 & 96 & \textbf{100} & 56.1 & 96 & \textbf{100} \\
        \midrule
        TransFuser~\cite{Nguyen2026CVPR}           & 46.6 & \underline{99} & 75 & 50.7 & 97 & 84 \\
        \quad Pisco                                & \textbf{59.8} & 90 & 40 & \underline{58.2} & 92 & 34 \\ %
        \quad Pisco Sour                           & \underline{59.6} & 96 & 64 & \textbf{64.1} & \underline{98} & 56 \\ %
        \bottomrule
    \end{tabular*}
    \vspace{-0.1cm}
    \caption{\textbf{Closed-loop results on Odyssey~\cite{Kim2026ARXIV}.} Route driving score (RDS), SD-route compliance (SDC) and comfort (C) in non-reactive and reactive traffic. Baseline results are taken from~\cite{Kim2026ARXIV}.}
    \label{tab:odyssey_internal}
    \vspace{-0.5cm}
\end{table}
\section{Conclusion}
\label{sec:conclusion}

We present Pisco Sour, a post-training recipe for end-to-end driving. Pisco distills a privileged RL teacher into the student, while Sour mixes human demonstrations into the distillation. The same recipe improves a 68M-parameter TransFuser and the 10B-parameter FlashDrive, achieving a new state of the art in AlpaSim. Pisco Sour with additional nuPlan scenes also improves TransFuser on HUGSIM and Odyssey. Our experiments show that the student inherits the limitations of the teacher with Pisco alone, and that human demonstrations through Sour can counteract these limitations while keeping the achieved improvements.

\boldparagraph{Limitations} In our experiments, the teacher unroll in Pisco supervises only the first 2\,s of the student's 6.4\,s long output trajectories, with longer horizon distillation necessitating additional computational overhead. It also introduces an asymmetry between the teacher and student, wherein the teacher observes the scene's logged future. Distilling in the action space of the teacher would avoid the unroll, but restricts the approach to students that output the same low-dimensional actions, which is uncommon for existing end-to-end models. We leave these explorations to future work.

\section*{LLM Usage Statement}
We used Claude (Anthropic) and Codex (OpenAI) as coding assistants to write and debug training, evaluation and analysis code, to write scripts and monitor experiments, and to write plotting scripts for figures and tables.
We also used Claude to draft and edit parts of the text and to check grammar.
The authors designed the method and experiments, reviewed the code, verified all results and text, and selected all citations.
The authors take full responsibility for the content of this paper.

{
    \small
    \bibliographystyle{ieeenat_fullname}
    \bibliography{bib/bibliography_long,bib/bibliography,bib/bibliography_custom}
}

\end{document}